\documentclass[letterpaper]{article} 
\usepackage[preprint]{aaai2027}
\usepackage[hyphens]{url}  
\usepackage{graphicx} 
\usepackage{natbib}  
\usepackage{caption} 
\usepackage{algorithm}
\usepackage{algorithmic}
\usepackage{newfloat}
\usepackage{listings}
\usepackage{booktabs}
\usepackage{colortbl}
\usepackage{amsmath}
\usepackage{amssymb}
\usepackage{marvosym}

\DeclareCaptionStyle{ruled}{labelfont=normalfont,labelsep=colon,strut=off} 
\floatstyle{ruled}
\newfloat{listing}{tb}{lst}{}
\floatname{listing}{Listing}

\title{Group Contrastive Policy Optimization: Learning Beyond the Self-Exploration Limit with Golden Answers}

\author{
    Hao Wu\textsuperscript{*},
    Wei Liu\textsuperscript{*},
    Shizhe Xiang\textsuperscript{*},
    Jinwen Luo,
    Yeshani,
    Jin Ma\textsuperscript{\Letter}
}

\affiliations{
    Tencent\\
    \{howu, gyroliu, shizhexiang, jamsluo, lisaniye, daniellwang\}@tencent.com
}

\begin{document}

\maketitle
\begingroup
\renewcommand{\thefootnote}{*}
\footnotetext[1]{Equal contribution.}
\renewcommand{\thefootnote}{\Letter}
\footnotetext[2]{Corresponding author.}
\endgroup

\begin{abstract}
Reinforcement learning with verifiable rewards (RLVR) has become a powerful paradigm for improving the reasoning abilities of large language models, yet group-based methods lose their learning signal when every sampled response receives zero reward. We propose Group Contrastive Policy Optimization (GCPO), a hybrid objective that preserves standard on-policy optimization for informative rollout groups and activates a proximal contrastive objective only when exploration completely fails. The auxiliary objective pairs an offline golden reasoning trajectory with the current policy's failed responses, thereby providing a reliable positive reference and prompt-specific hard negatives without treating the golden trajectory as an on-policy rollout. Its old-policy token probabilities are evaluated through teacher forcing and used solely to constrain the update, rather than as an importance-sampling correction for the teacher distribution. A first-order analysis shows that the resulting update increases a length-normalized likelihood margin between reliable reasoning and observed failures. Experiments across two model scales and 10 core mathematical reasoning benchmarks demonstrate that GCPO improves training efficiency and final reasoning performance over competitive supervised and RLVR baselines.

\end{abstract}

\section{Introduction}

Recent advances in large language models have shown that reinforcement learning with verifiable rewards (RLVR) can serve as an effective post-training paradigm for enhancing reasoning capabilities\cite{openai2024openai,guo2025deepseek,cobbe2021training}. In particular, the success of models such as OpenAI-o1 and DeepSeek-R1 has shifted increasing attention from purely supervised fine-tuning on curated reasoning traces toward reinforcement-based training, where the model is encouraged to discover effective reasoning strategies through interaction with verifiable feedback\cite{openai2024openai,shao2024deepseekmath}. Unlike supervised fine-tuning, which imitates ground-truth solutions under teacher forcing\cite{bengio2015scheduled}, RLVR uses outcome-level rewards to explore diverse reasoning trajectories and selectively reinforce correct answers\cite{cobbe2021training,shao2024deepseekmath,lightman2024let}. As supported by prior studies~\cite{lightman2024let,luong2024reft}, this exploration-driven process is crucial for reasoning generalization because it enables problem-solving behaviors beyond fixed demonstrations and supports adaptation to novel problem structures.

\begin{figure*}[t]
    \centering
    \begin{minipage}[t]{0.485\textwidth}
        \centering
        \includegraphics[width=\linewidth]{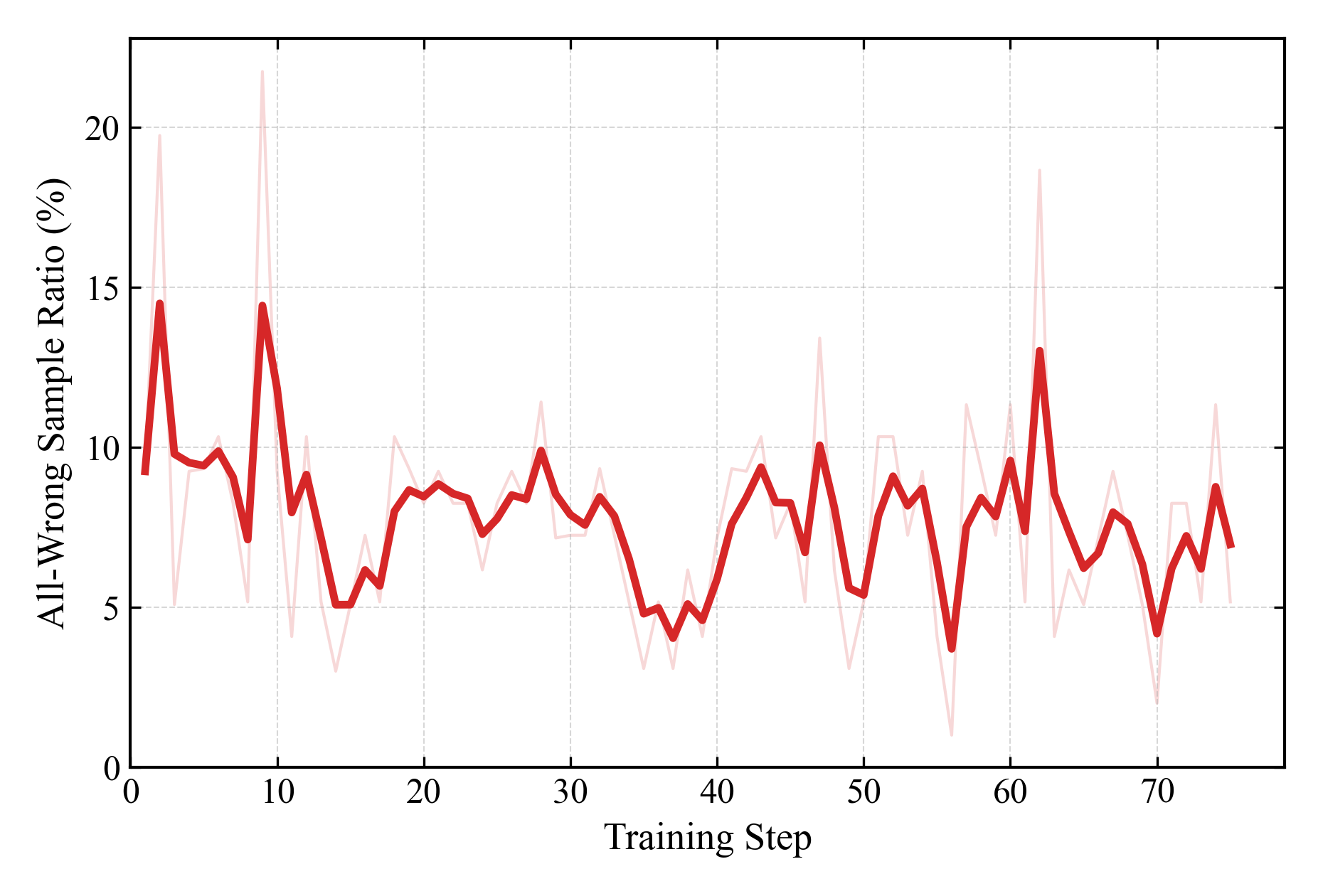}
        \textbf{(a)} All-wrong sample ratio during training.
    \end{minipage}
    \hfill
    \begin{minipage}[t]{0.485\textwidth}
        \centering
        \includegraphics[width=\linewidth]{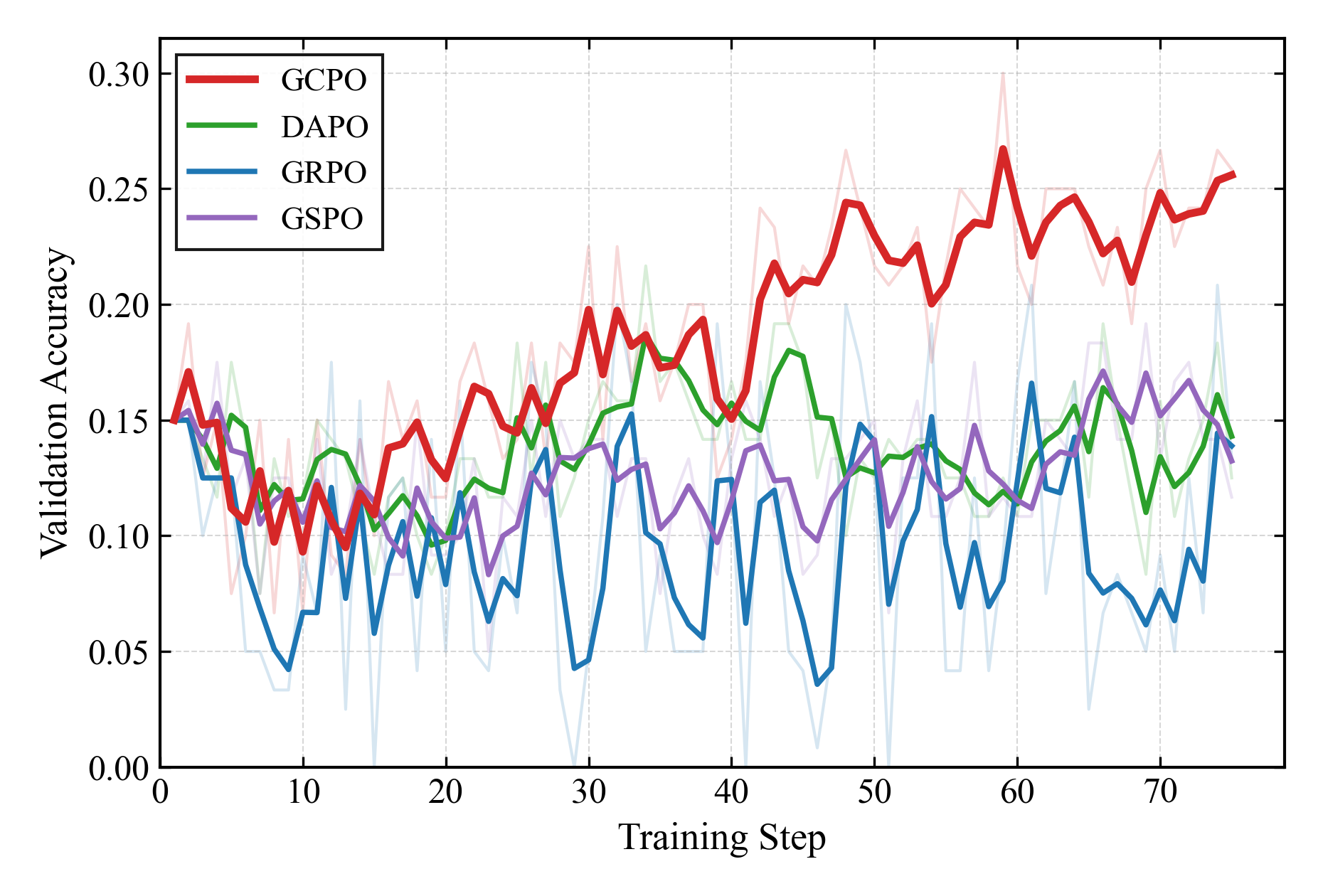}
        \textbf{(b)} Validation accuracy during training.
    \end{minipage}
    \caption{Training dynamics of GCPO. (a) Ratio of all-wrong rollout groups. (b) Validation accuracy of GCPO and RLVR baselines. Solid lines show smoothed trends; lighter lines show raw values.}
    \label{fig:intro_motivation}
\end{figure*}

Existing RLVR methods for mathematical reasoning, such as Group Relative Policy Optimization (GRPO) and its variants \cite{shao2024deepseekmath,yu2026dapo,zheng2025group}, commonly rely on group-wise relative advantages to construct learning signals. For each problem, the policy samples a group of candidate responses, which are evaluated by a rule-based verifier and normalized within the group. This relative formulation allows the model to distinguish better responses from worse ones under the same problem context, thereby providing an effective signal for learning what constitutes a higher-quality reasoning trajectory.

However, this relative formulation comes with an implicit prerequisite: the sampled group must contain sufficient reward variation so that a meaningful comparison can be constructed. In practice, this assumption often breaks down during the early stage of RLVR. To quantify the prevalence of this failure mode, we track the proportion of all-wrong groups throughout training. As shown in Figure~\ref{fig:intro_motivation}(a), such groups persist across the entire training process and account for a non-negligible fraction of the sampled prompts. For these prompts, group-wise reward normalization produces no informative advantage, leaving the policy with little usable learning signal on exactly the problems that require the most improvement. As a result, the training process becomes bounded by the policy’s own exploration ability: difficult samples are either under-utilized or discarded, and the model can only improve within the region that it is already able to explore. A straightforward workaround is to fine-tune the model on ground-truth solutions, but teacher-forced imitation lacks contrastive feedback and generalizes poorly to challenging, previously unseen reasoning problems.

To overcome this exploration bottleneck, we propose Group Contrastive Policy Optimization (GCPO), which augments on-policy GRPO with a failure-triggered proximal contrastive objective. When all sampled responses receive zero rewards, GCPO pairs an offline golden reasoning trajectory with the remaining on-policy failures. The golden trajectory is a supervised positive reference rather than a rollout from the old policy, while the failures retain the model's current error modes as hard negatives. This distinction yields a clean hybrid objective: ordinary rollout groups continue to use standard RLVR, whereas all-wrong groups receive an auxiliary contrastive update that moves the policy toward verified reasoning and away from its observed failures. The validation curves in Figure~\ref{fig:intro_motivation}(b) provide an early empirical view of this benefit: GCPO increasingly separates from competitive group-based RLVR baselines as training proceeds. Together with the persistent all-wrong ratio in Figure~\ref{fig:intro_motivation}(a), this result indicates that targeted contrastive supervision can improve training efficiency and extend the model's effective exploration boundary. We analyze the local optimization behavior of this objective and validate it on diverse mathematical reasoning benchmarks.

In summary, this paper makes the following contributions:

\begin{itemize}
    \item We construct a high-quality golden reasoning trajectory dataset for failure-aware post-training, pairing each prompt with a verified reference solution containing both the correct answer and a complete reasoning process.

    \item We propose GCPO, a hybrid objective that combines on-policy GRPO with a proximal contrastive term activated only for all-wrong groups. It couples an offline positive reference with online hard negatives without interpreting the reference as an on-policy sample.

    \item Experiments on two model scales and 10 core benchmarks show that GCPO improves training efficiency and reasoning performance over competitive supervised and RLVR baselines.
\end{itemize}

\section{Related Work}

\begin{figure*}[t]
    \centering
    \includegraphics[width=\textwidth]{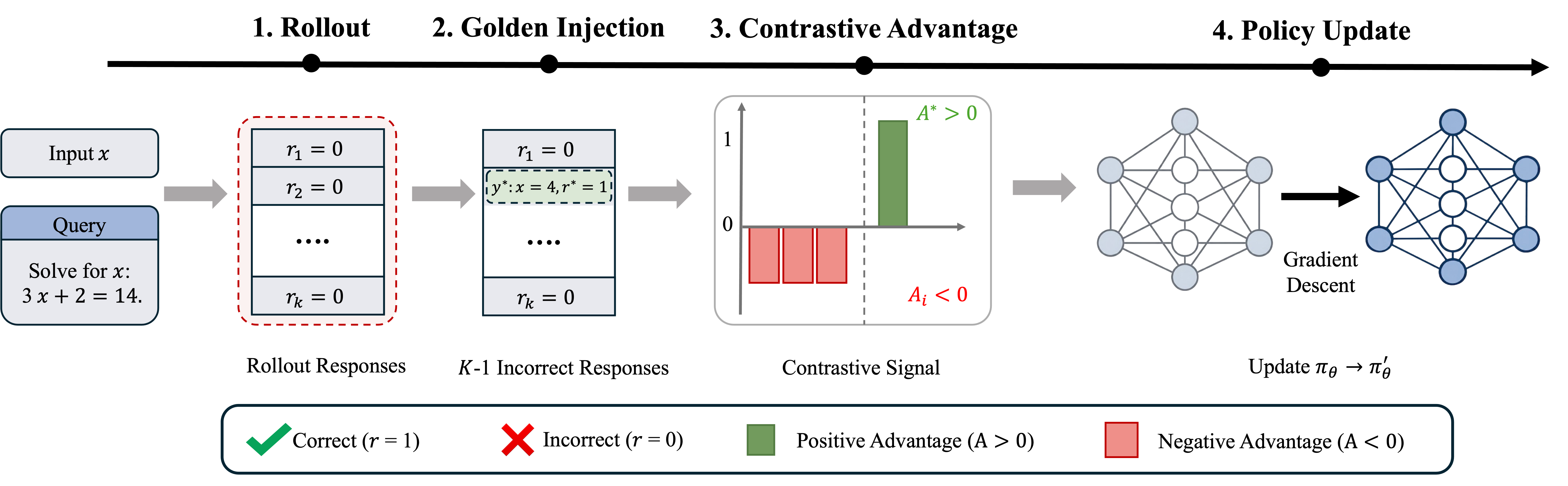}
    \caption{Overview of GCPO. An all-wrong group activates a proximal contrastive objective between an offline golden reference and online failures.}
    \label{fig:gcpo_overview}
\end{figure*}

\paragraph{Reinforcement Learning with Verifiable Rewards for Reasoning.}
Reinforcement learning has become a central post-training paradigm for improving the reasoning capabilities of large language models\cite{guo2025deepseek}. Instead of relying solely on supervised imitation of curated solutions, reinforcement learning with verifiable rewards (RLVR) optimizes policies using outcome-level feedback provided by rule-based verifiers, making it particularly suitable for domains such as mathematical reasoning and code generation\cite{le2022coderl}. Early RLHF methods typically employ PPO with learned reward models\cite{ouyang2022training}, while recent reasoning-oriented methods, such as GRPO, remove the value model by estimating advantages from groups of sampled responses\cite{shao2024deepseekmath}. Building upon this formulation, subsequent variants further improve training stability, sample efficiency, and reward utilization\cite{yu2026dapo}. However, these methods largely depend on the policy's own rollouts to construct informative learning signals\cite{zheng2025group}. When the model fails to generate any correct response for a difficult prompt, group-wise relative advantages collapse, leaving the policy without a clear optimization direction. This makes existing RLVR methods inherently constrained by the model's self-exploration capability.

\paragraph{Learning from Reasoning Trajectories and Preference Signals.}
Another line of work improves reasoning by training models on high-quality chain-of-thought trajectories, either through supervised fine-tuning or distillation from stronger teacher models\cite{wei2022chain,hsieh2023distilling}. Such methods transfer explicit problem-solving patterns but mainly perform positive imitation under teacher forcing, without contrasting a reference against the learner's own errors\cite{bengio2015scheduled}. Preference-based methods such as DPO and RLHF instead distinguish preferred from rejected responses\cite{rafailov2023direct,ouyang2022training}. Closely related to our work, LUFFY treats external demonstrations as off-policy samples participating in mixed-policy RL and introduces regularized importance sampling to balance imitation and exploration~\cite{yan2025learning}. GCPO takes a different route: it does not use the golden trajectory to estimate the on-policy RL return. Instead, the trajectory acts as an empirical positive in a separate proximal contrastive objective that is activated only when all on-policy responses fail. GCPO thus constructs a failure-conditioned preference instance from an offline positive and online hard negatives.

\section{Methodology}

\subsection{Preliminaries}

\paragraph{Supervised Fine-Tuning.}
Given a prompt $x$ and a golden solution $y^{\star}$, supervised fine-tuning (SFT) minimizes
\begin{equation}
    \mathcal{L}_{\mathrm{SFT}}(\theta)
    =
    - \mathbb{E}_{(x,y^{\star}) \sim \mathcal{D}}
    \left[
        \log \pi_{\theta}(y^{\star} \mid x)
    \right].
\end{equation}
SFT provides direct supervision but does not explicitly contrast the reference with errors produced by the current policy.

\paragraph{Direct Preference Optimization.}
Given a preferred response $y^{+}$ and a rejected response $y^{-}$, Direct Preference Optimization (DPO) increases their reference-normalized likelihood margin:
\begin{equation}
\begin{aligned}
    \Delta_\theta
    &=
    \log\frac{\pi_\theta(y^+\mid x)}{\pi_{\mathrm{ref}}(y^+\mid x)}
    -
    \log\frac{\pi_\theta(y^-\mid x)}{\pi_{\mathrm{ref}}(y^-\mid x)},\\
    \mathcal L_{\mathrm{DPO}}
    &=-\mathbb E_{\mathcal D}\!\left[\log\sigma(\beta\Delta_\theta)\right].
\end{aligned}
\end{equation}
Here, $\pi_{\mathrm{ref}}$ is fixed and $\beta$ controls the preference strength.

\paragraph{Group Relative Policy Optimization.}
For each prompt $x$, GRPO samples $\{y_i\}_{i=1}^{G}$ from the old policy $\pi_{\theta_{\mathrm{old}}}$ and obtains verifier rewards $\{r_i\}_{i=1}^{G}$~\cite{schulman2017proximal,shao2024deepseekmath}. Let $\hat A_i$ denote the group-normalized advantage. The token-level clipped objective is
\begin{equation}
\begin{aligned}
    \mathcal{J}_{\mathrm{GRPO}}(\theta)
    =
    \frac{1}{G}
    \sum_{i=1}^{G}
    \frac{1}{|y_i|}
    \sum_{t=1}^{|y_i|}
    \min
    \left(
        \rho_{i,t}\hat{A}_i,
        \bar{\rho}_{i,t}\hat{A}_i
    \right),
\end{aligned}
\end{equation}
where
\begin{equation}
\begin{aligned}
    \rho_{i,t}
    &=
    \frac{\pi_\theta(y_{i,t}\mid x,y_{i,<t})}
         {\pi_{\theta_{\mathrm{old}}}(y_{i,t}\mid x,y_{i,<t})},\\
    \bar\rho_{i,t}
    &=\operatorname{clip}
    \left(\rho_{i,t},1-\epsilon_{\mathrm{low}},
    1+\epsilon_{\mathrm{high}}\right).
\end{aligned}
\end{equation}
The sampling measure is important: the standard surrogate assumes that every $y_i$ is generated by $\pi_{\theta_{\mathrm{old}}}$. When all rewards are zero, group normalization yields $\hat A_i=0$ and the objective provides no update.

\begin{figure*}[t]
    \centering
    \includegraphics[width=0.74\textwidth]{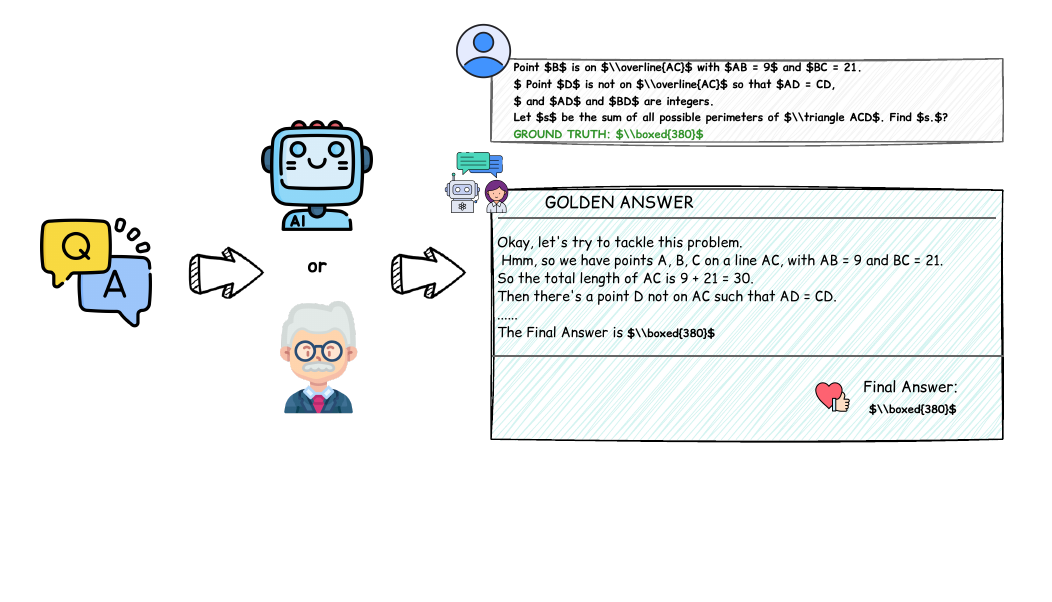}
    \vspace{-40 pt}
    \caption{Golden trajectory construction using a strong teacher model or human expert.}
    
    \label{fig:golden_construction}
\end{figure*}

\paragraph{Golden Trajectory Construction.}
Before training, we construct an offline golden trajectory $y^{\star}$ for each prompt using DeepSeek-R1 as the teacher model~\cite{guo2025deepseek}, selected for its strong performance on complex reasoning tasks. We sample up to 32 candidate trajectories per prompt and retain only answer-consistent, complete reasoning traces; prompts for which no verified correct trajectory is found are treated as extremely difficult and excluded. We also manually inspect a subset of the retained groups to verify the correctness of their intermediate reasoning rather than relying solely on final-answer matching. As illustrated in Figure~\ref{fig:golden_construction}, the retained $y^\star$ is drawn from an external data distribution rather than from $\pi_{\theta_{\mathrm{old}}}$.

\subsection{Group Contrastive Policy Optimization}

\paragraph{Failure-Triggered Contrastive Group.}
Let $\mathcal E_{\mathrm{fail}}$ denote the event that all $G$ on-policy responses fail verification:
\begin{equation}
    \mathcal E_{\mathrm{fail}}
    =\{r_1=\cdots=r_G=0\},
    \qquad \hat A_1=\cdots=\hat A_G=0 .
\end{equation}
The trigger is determined from the original on-policy group before any modification. Under this event, GCPO forms an auxiliary contrastive group containing the offline reference $y^\star$ and $G-1$ online failures $\{y_i^-\}_{i=1}^{G-1}$. For implementation, this is realized by replacing one zero-reward response and assigning pseudo-rewards
\begin{equation}
    \tilde r(y^{\star}) = 1,
    \qquad
    \tilde r(y_i^{-}) = 0 .
\end{equation}
Omitting the numerical stabilizer, which only induces a common positive
rescaling, group normalization gives the exact weights
\begin{equation}
\label{eq:contrastive_weights}
    \hat{A}^{\star}
    =
    \sqrt{G-1},
    \qquad
    \hat{A}^{-}_i
    =
    -\frac{1}{\sqrt{G-1}} .
\end{equation}

\paragraph{Hybrid Proximal Objective.}
GCPO uses standard GRPO whenever the rollout group is not all wrong. If every
rollout fails, the GRPO advantages are zero, so GCPO instead uses the proximal
contrastive objective. The switch is
\begin{equation}
\mathcal J_{\mathrm{GCPO}}=
\begin{cases}
    \mathcal J_{\mathrm{PC}},
        & \mathcal E_{\mathrm{fail}},\\
    \mathcal J_{\mathrm{GRPO}},
        & \text{otherwise},
\end{cases}
\end{equation}
For a response $y$ with weight $A$, define
\begin{equation}
\begin{aligned}
    \ell_{\mathrm{prox}}(y,A)
    &=
    \frac{1}{|y|}
    \sum_{t=1}^{|y|}
    \min\!\left(\rho_t(y)A,\bar\rho_t(y)A\right),\\
    \mathcal J_{\mathrm{PC}}
    &=
    \frac{1}{G}
    \left[
    \ell_{\mathrm{prox}}(y^\star,\hat A^\star)
    \right.\\[-2pt]
    &\hspace{31mm}\left.
    +\sum_{i=1}^{G-1}
    \ell_{\mathrm{prox}}(y_i^-,\hat A_i^-)
    \right].
\end{aligned}
\end{equation}
For on-policy failures, old log-probabilities are stored during rollout. For $y^\star$, they are evaluated token by token under the same frozen old-policy snapshot through teacher forcing. The ratio $\rho_t(y^\star)$ is therefore a \emph{proximal reference ratio}: clipping it limits the magnitude of a policy update, but does not correct the fact that $y^\star$ originates from an external teacher distribution. GCPO consequently does not claim that $\mathcal J_{\mathrm{PC}}$ is an unbiased on-policy policy-gradient estimator; it is a well-defined empirical auxiliary objective over an offline positive and online negatives. Because it does not use $y^\star$ to estimate the on-policy return, it requires no inaccessible teacher-density importance ratio.

\paragraph{First-Order Contrastive Interpretation.}
Define the length-normalized response score
\begin{equation}
    s_\theta(y\mid x)
    =
    \frac{1}{|y|}
    \sum_{t=1}^{|y|}
    \log\pi_\theta(y_t\mid x,y_{<t}) .
\end{equation}
Near the start of an update, $\theta\approx\theta_{\mathrm{old}}$, all ratios are near one, and clipping is inactive. The local update direction is then
\begin{equation}
\begin{aligned}
    \nabla_\theta\mathcal J_{\mathrm{PC}}
    &\propto
    \hat A^\star\nabla_\theta s_\theta(y^\star\mid x)
    +\sum_{i=1}^{G-1}\hat A_i^-
    \nabla_\theta s_\theta(y_i^-\mid x)\\
    &\propto
    \nabla_\theta
    \left[
    s_\theta(y^\star\mid x)
    -\frac{1}{G-1}\sum_{i=1}^{G-1}s_\theta(y_i^-\mid x)
    \right].
\end{aligned}
\end{equation}
This first-order result explains the optimization behavior rather than asserting off-policy unbiasedness: GCPO increases the normalized likelihood of verified reasoning relative to failures sampled from the policy's current weaknesses. Unlike SFT, it uses explicit negative feedback; unlike DPO, its negatives are refreshed online and the objective is activated only after complete exploration failure.

\begin{table*}[t]
    \centering
    \caption{Pass@1 accuracy (\%) on 10 mathematical reasoning benchmarks. Avg. denotes the mean over all reported benchmarks. The best result within each model scale is shown in bold.}
    \label{tab:main_results}
    \renewcommand{\arraystretch}{1.08}
    \resizebox{\textwidth}{!}{%
    \begin{tabular}{l|cccccccccc|c}
        \toprule
        Method &
        GSM8K &
        \shortstack{Minerva\\Math} &
        SVAMP &
        ASDiv &
        MAWPS &
        TabMWP &
        \shortstack{MMLU\\STEM} &
        \shortstack{AIME\\2024} &
        \shortstack{AIME\\2025} &
        \shortstack{Olympiad\\Bench} &
        Avg. \\
        \midrule
        \multicolumn{12}{l}{\textit{DeepSeek-R1-Distill-Qwen-1.5B}} \\
        \quad Base    & 43.30 & 34.60 & 57.00 & 59.90 & 67.90 & 41.50 & 10.90 & 17.81 & 19.21 & 36.37 & 38.85 \\
        \quad + SFT   & 30.60 & 33.80 & 31.00 & 31.00 & 33.80 & 31.70 & 27.00 & 16.46 & 19.58 & 37.48 & 29.24 \\
        \quad + GRPO  & 40.80 & 43.40 & 45.20 & 49.20 & 59.70 & 31.20 & 24.90 & 13.33 & 20.92 & 39.35 & 36.80 \\
        \quad + DAPO  & 51.60 & 34.20 & 65.90 & 67.70 & 74.20 & 50.00 & 14.90 & 18.44 & 23.37 & \textbf{46.67} & 44.70 \\
        \quad + GSPO  & 53.80 & \textbf{55.40} & 53.00 & 54.10 & 63.80 & 44.40 & \textbf{32.70} & 18.02 & 20.47 & 40.67 & 43.64 \\
        \quad + LUFFY & 34.70 & 45.00 & 37.30 & 39.90 & 50.50 & 27.40 & 25.20 & 15.83 & 21.28 & 41.87 & 33.90 \\
        \rowcolor{black!10}
        \quad + GCPO  & \textbf{56.50} & 40.40 & \textbf{67.90} & \textbf{70.90} & \textbf{76.30} & \textbf{55.30} & 18.70 & \textbf{19.58} & \textbf{23.68} & \textbf{46.67} & \textbf{47.59} \\
        \midrule
        \multicolumn{12}{l}{\textit{DeepSeek-R1-Distill-Llama-8B}} \\
        \quad Base   & 86.70 & 88.20 & 88.00 & 88.80 & 93.10 & 85.70 & 35.90 & 26.25 & 20.10 & 41.54 & 65.43 \\
        \quad + SFT  & 79.70 & 81.40 & 84.70 & 83.30 & 89.00 & \textbf{86.70} & \textbf{47.80} & 25.00 & 21.65 & 42.70 & 64.20 \\
        \quad + DAPO & 87.40 & 87.80 & 89.10 & 89.10 & 94.40 & 86.10 & 36.60 & 28.33 & 20.89 & 43.44 & 66.32 \\
        \quad + GSPO & 87.30 & 87.60 & 89.30 & 89.70 & 93.20 & 85.30 & 35.10 & 28.75 & 20.29 & 41.69 & 65.82 \\
        \quad + LUFFY & 87.70 & 86.40 & \textbf{89.70} & 89.70 & 92.10 & 82.80 & 47.70 & 25.00 & 21.03 & 42.59 & 66.47 \\
        \rowcolor{black!10}
        \quad + GCPO & \textbf{88.60} & \textbf{89.00} & 88.90 & \textbf{89.80} & \textbf{94.60} & 85.90 & 37.00 & \textbf{30.00} & \textbf{21.84} & \textbf{44.50} & \textbf{67.01} \\
        \bottomrule
    \end{tabular}%
    }
\end{table*}

\subsection{Why GCPO Works}

The hybrid-objective view explains why a small auxiliary intervention is useful exactly when group-based RLVR fails: the proximal contrastive term supplies a prompt-conditioned comparison between verified reasoning and the policy's current failures.

\paragraph{Recovering Directional Gradients.}
Identical zero rewards erase every on-policy advantage. Activating $\mathcal J_{\mathrm{PC}}$ breaks this symmetry by assigning a positive weight to the offline reference and negative weights to failed responses, restoring a directional update without relabeling the reference as an on-policy sample.

\paragraph{Concentrating Supervision on Difficult Prompts.}
All-wrong groups identify prompts beyond the policy's current exploration capability. GCPO adds external supervision only in these zero-gradient cases and leaves standard on-policy learning unchanged whenever the model discovers reward variation.

\paragraph{Contrastive Distillation of Reasoning.}
A complete golden trajectory transfers structured reasoning rather than only a final answer. Pairing this offline positive with freshly sampled failures simultaneously distills verified reasoning and suppresses current error modes, providing richer feedback than positive-only teacher forcing.

\paragraph{Proximal Stabilization.}
Teacher-forced old-policy probabilities define a local anchor for the offline reference. Token-level clipping bounds abrupt changes to both positive and negative responses, while ordinary reward differences continue to govern non-degenerate groups. This is a stabilization mechanism for the empirical auxiliary objective, not an importance correction for the teacher distribution or a general convergence guarantee.

Overall, GCPO augments failed exploration with a dynamically constructed preference signal whose positive is offline but whose hard negatives track the policy's current weaknesses.

\begin{figure*}[t]
    \centering
    \begin{minipage}[t]{0.48\textwidth}
        \centering
        \includegraphics[width=\linewidth]{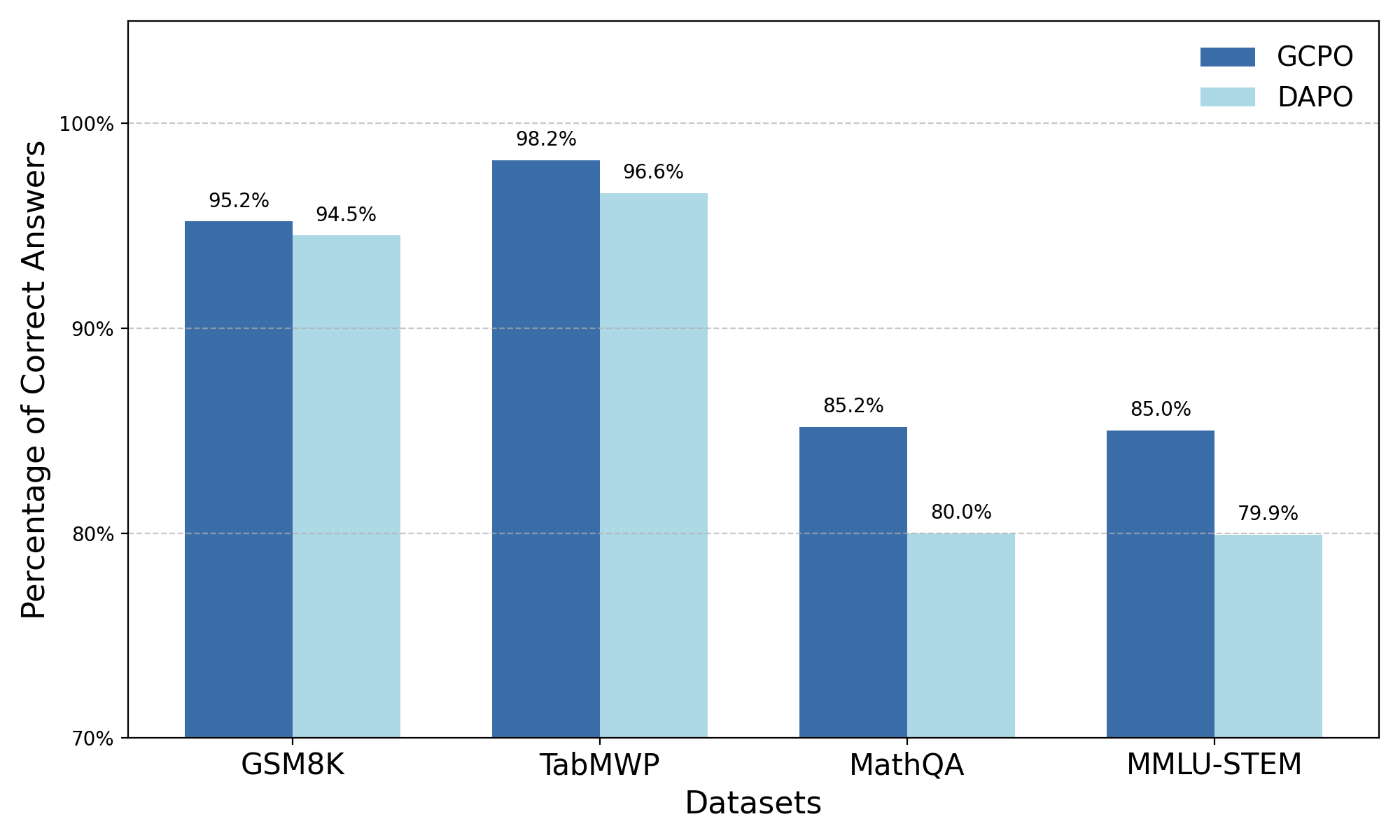}
        \small (a) Pass@32
    \end{minipage}
    \hfill
    \begin{minipage}[t]{0.48\textwidth}
        \centering
        \includegraphics[width=\linewidth]{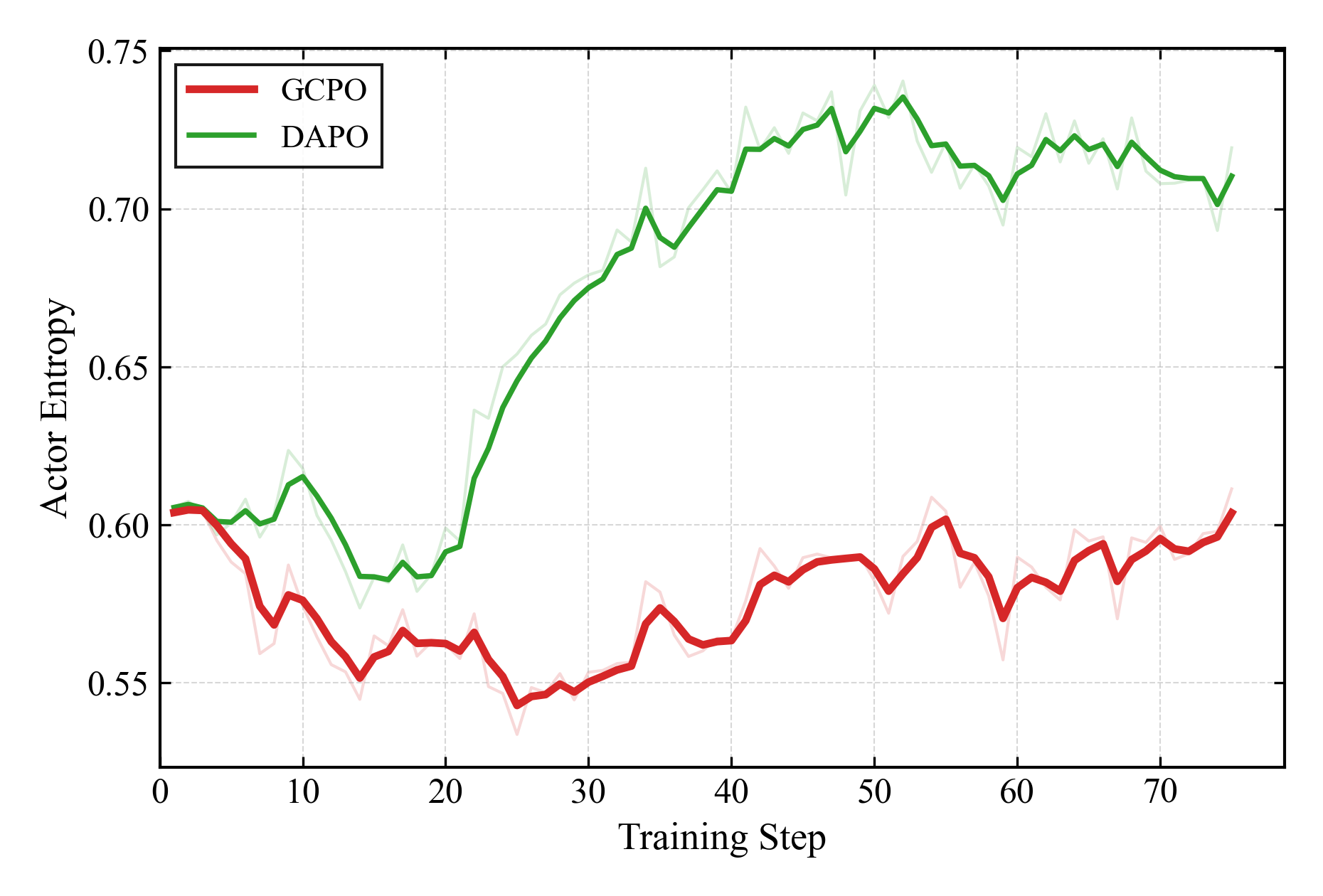}
        \small (b) Actor entropy
    \end{minipage}
    \caption{Answer quality and policy dynamics of GCPO and DAPO.}
    \label{fig:quality_policy_dynamics}
\end{figure*}

\section{Experiments}

\subsection{Experimental Setup}

\looseness=-1
We evaluate GCPO using two reasoning models with different scales and architectures: DeepSeek-R1-Distill-Qwen-1.5B and DeepSeek-R1-Distill-Llama-8B~\cite{guo2025deepseek}. Our main evaluation covers 10 mathematical reasoning benchmarks: AIME 2024 and AIME 2025~\cite{aime2024}, GSM8K~\cite{cobbe2021training}, Minerva Math~\cite{lewkowycz2022solving}, ASDiv~\cite{miao2020diverse}, MAWPS~\cite{koncel2016mawps}, TabMWP~\cite{lu2023dynamic}, MMLU-STEM~\cite{hendrycks2021measuring}, SVAMP~\cite{patel2021nlp}, and OlympiadBench~\cite{he2024olympiadbench}. MathQA~\cite{amini2019mathqa} is used only in the separate answer-quality diagnostic. Following the original evaluation protocol, we report pass@1 accuracy averaged over 8 independent rollouts per problem to reduce the variance caused by stochastic decoding. The sampling temperature is set to 0.7 for both training and evaluation.

\looseness=-1
For training, we use DAPO-Math-17k-Processed from Open-R1~\cite{huggingface2025openr1} and employ DeepSeek-R1 to generate one complete golden reasoning trajectory for each problem. After filtering, the resulting training set contains 9,975 examples. We implement all methods with verl~\cite{sheng2024hybridflow} and train the 1.5B and 8B models on 8 and 32 NVIDIA H20 GPUs, respectively, with 16 rollouts per prompt and a maximum sequence length of 16k tokens. The failure trigger is computed from the original on-policy group. If all responses are incorrect, the implementation replaces the first zero-reward response to form the auxiliary contrastive group. For this offline reference, the frozen old policy computes token-level log-probabilities through teacher forcing; ratios are clipped token by token and aggregated over valid response tokens. Thus, these ratios constrain the auxiliary update and are not teacher-distribution importance weights. All remaining hyperparameters follow the default DAPO configuration~\cite{yu2026dapo}.

\begingroup
\setlength{\intextsep}{4pt}
\begin{table}[H]
    \centering
    \captionsetup{skip=3pt}
    \caption{Training hyperparameters shared by all methods.}
    \label{tab:training_hyperparameters}
    \renewcommand{\arraystretch}{1.00}
    \resizebox{\columnwidth}{!}{%
    \begin{tabular}{lc|lc}
        \toprule
        Config & Setting & Config & Setting \\
        \midrule
        Optimizer & AdamW & Training epochs & 3 \\
        Training batch size & 128 & Mini-batch size & 64 \\
        Learning rate & $1.0\times10^{-6}$ & Warmup ratio & 0.1 \\
        Generations per prompt & 16 & Max prompt length & 2,048 \\
        Max completion length & 14,336 & Clip ratio low & 0.20 \\
        Clip ratio high & 0.28 & Top-$p$ & 0.95 \\
        Top-$k$ & 20 & Temperature & 0.7 \\
        \bottomrule
    \end{tabular}%
    }
\end{table}
\endgroup

\looseness=-1
Table~\ref{tab:training_hyperparameters} reports the configuration used across methods and model scales. We optimize with AdamW for three epochs using a training batch size of 128, a mini-batch size of 64, a learning rate of $1.0\times10^{-6}$, and a 10\% warmup ratio. Holding these optimization parameters fixed ensures that performance differences are attributable to the learning objective rather than method-specific tuning.

\looseness=-1
Each prompt produces 16 candidate completions. The 2,048-token prompt limit and 14,336-token completion limit together form the 16,384-token context budget; golden trajectories are retained only when they remain complete within this budget. We use asymmetric clipping with lower and upper ratios of 0.20 and 0.28, respectively. Rollouts are sampled with top-$p=0.95$, top-$k=20$, and temperature $0.7$, providing sufficient response diversity while keeping the training and evaluation protocol consistent.

\begingroup
\setlength{\parskip}{0pt}
\subsection{Main Results}

\looseness=-1
Table \ref{tab:main_results} compares GCPO with supervised fine-tuning and representative RLVR baselines across model scales.

\paragraph{Results on the 1.5B Model.}

\looseness=-1
GCPO achieves the strongest average performance, reaching 47.59\% across the 10 reported benchmarks, compared with 44.70\% for DAPO and 43.64\% for GSPO. It obtains the best or tied-best result on eight benchmarks and improves the base model by 8.74 percentage points on average. The gains are particularly clear on TabMWP, ASDiv, GSM8K, MAWPS, and SVAMP, where GCPO surpasses the strongest competing method by 5.30, 3.20, 2.70, 2.10, and 2.00 points, respectively. This pattern is consistent with GCPO's failure-triggered design: lower-success prompts expose more all-wrong groups and therefore more opportunities for targeted contrastive supervision in the smaller model.

\paragraph{Results on the 8B Model.}
\looseness=-1
At the larger scale, GCPO achieves the highest 10-benchmark average of 67.01\%, outperforming LUFFY, DAPO, and GSPO by 0.54, 0.69, and 1.19 percentage points, respectively. It attains the best result among all methods on seven benchmarks and surpasses LUFFY on eight of the 10 tasks. Its advantages are most evident on competition-level evaluations: GCPO reaches 30.00\%, 21.84\%, and 44.50\% on AIME 2024, AIME 2025, and OlympiadBench, improving over LUFFY by 5.00, 0.81, and 1.91 points. It also achieves the highest scores on GSM8K, Minerva Math, ASDiv, and MAWPS, while LUFFY remains stronger on SVAMP and MMLU-STEM. These results show that GCPO transfers effectively across scales and remains especially competitive on difficult mathematical reasoning tasks.
\endgroup

\begingroup
\setlength{\intextsep}{4pt}
\begin{table}[H]
    \centering
    \captionsetup{skip=3pt}
    \caption{Ablation results on the 1.5B model.}
    \label{tab:ablation}
    \renewcommand{\arraystretch}{1.03}
    \resizebox{\columnwidth}{!}{%
    \begin{tabular}{l|cccc}
        \toprule
        Method &
        \shortstack{MMLU\\STEM} &
        \shortstack{AIME\\2024} &
        \shortstack{AIME\\2025} &
        \shortstack{Olympiad\\Bench} \\
        \midrule
        Universal Injection & 9.80 & 9.90 & 20.36 & 38.45 \\
        Neg-Adv & 12.10 & 15.42 & 20.56 & 43.65 \\
        GCPO w/ KL & \textbf{20.20} & 18.13 & 22.32 & 45.35 \\
        \rowcolor{black!10}
        GCPO & 18.70 & \textbf{19.58} & \textbf{23.68} & \textbf{46.67} \\
        \bottomrule
    \end{tabular}%
    }
\end{table}
\endgroup

\begingroup
\setlength{\parskip}{0pt}
\subsection{Ablation Study}

\looseness=-1
We conduct ablations on DeepSeek-R1-Distill-Qwen-1.5B to examine the effects of activation scope, contrastive weighting and KL divergence. Table~\ref{tab:ablation} compares GCPO with three variants on four representative benchmarks.

\paragraph{Activation Scope.}
\looseness=-1
Universal Injection activates the golden-reference auxiliary objective for every group, including groups that already contain successful responses. This indiscriminate strategy consistently degrades performance, reducing accuracy by 9.68, 8.90, 3.32, and 8.22 percentage points on AIME 2024, MMLU-STEM, AIME 2025, and OlympiadBench, respectively. The result supports separating standard on-policy learning from external supervision and activating the latter only after complete exploration failure.

\paragraph{Advantage Assignment.}
\looseness=-1
Neg-Adv assigns every response in the auxiliary group the same negative weight used for failed rollouts, thereby removing the positive signal for the offline reference. It underperforms GCPO on all four benchmarks, showing that the gains do not arise merely from penalizing sampled responses; they require the contrast between a verified positive and on-policy failures.

\paragraph{KL Regularization.}
\looseness=-1
Adding fixed-reference KL improves MMLU-STEM by 1.50 points but lowers AIME 2024, AIME 2025, and OlympiadBench by 1.45, 1.36, and 1.32 points, reducing the average from 27.16\% to 26.50\%. This suggests that KL can restrict adaptation on challenging tasks; we therefore omit it.
\endgroup

\subsection{Answer Quality and Policy Dynamics}

We further compare 1.5B models on GCPO with DAPO using two complementary diagnostics: correct-answer rates on representative datasets and actor-entropy trajectories throughout training. Together, they connect GCPO's outcome-level gains with its underlying policy dynamics.

\paragraph{Correct-Answer Rate.}
As shown in Figure~\ref{fig:quality_policy_dynamics}(a), GCPO consistently produces a higher proportion of correct answers than DAPO. The gains are modest on GSM8K and TabMWP, where both methods already achieve high rates, with improvements of 0.7 and 1.6 percentage points, respectively. In contrast, GCPO yields substantially larger gains on MathQA and MMLU-STEM, improving over DAPO by 5.2 and 5.1 points. This pattern suggests that activating the auxiliary contrastive term on all-wrong groups is particularly beneficial on challenging tasks while preserving performance where on-policy exploration is already effective.

\paragraph{Actor Entropy.}
Figure~\ref{fig:quality_policy_dynamics}(b) reveals markedly different policy dynamics. DAPO's actor entropy rises sharply after approximately 20 training steps and remains above 0.70 during the later stage, whereas GCPO initially decreases to about 0.54 and subsequently stabilizes around 0.58--0.60. Lower entropy alone does not imply a better policy; however, its combination with higher accuracy suggests that the proximal contrastive signal concentrates probability on useful reasoning modes rather than inducing premature collapse. This coupling of lower entropy with higher accuracy distinguishes useful policy concentration from unproductive entropy reduction.

\begin{figure}[H]
    \centering
    \includegraphics[width=\columnwidth]{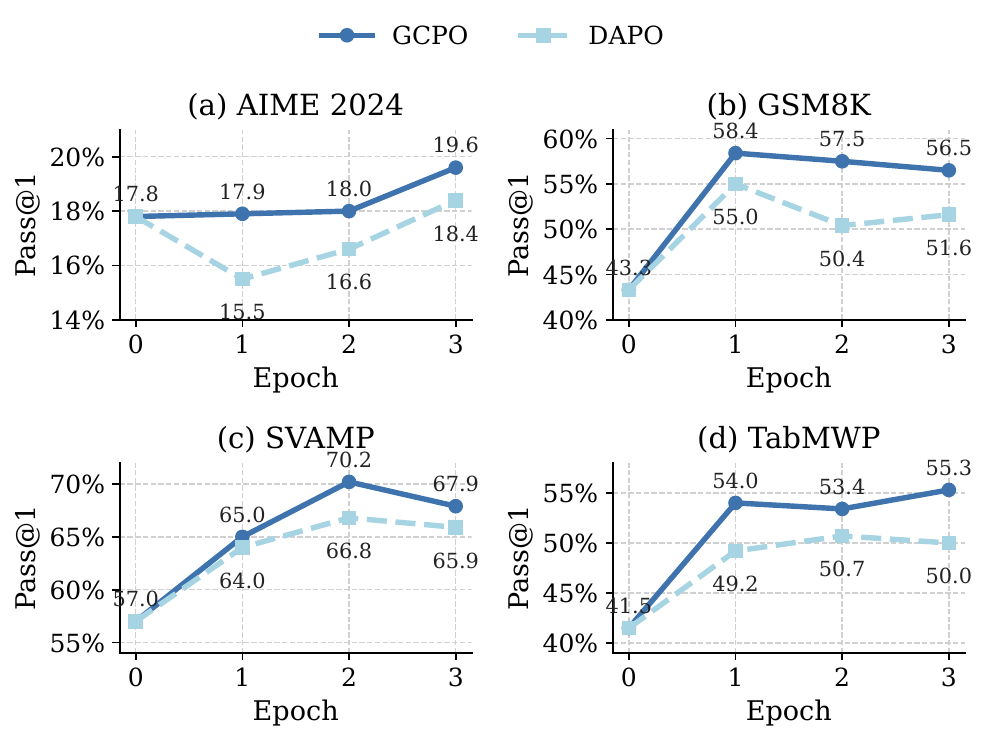}
    \caption{Training accuracy across epochs.}
    \label{fig:training_accuracy_dynamics}
\end{figure}

\subsection{Training Accuracy Dynamics}

Figure~\ref{fig:training_accuracy_dynamics} tracks pass@1 accuracy throughout training on four representative benchmarks. Starting from the same base checkpoint, GCPO establishes an advantage after only one epoch: it reaches 58.4\% versus 55.0\% on GSM8K, 65.0\% versus 64.0\% on SVAMP, and 54.0\% versus 49.2\% on TabMWP. On the more challenging AIME 2024 benchmark, GCPO maintains its initial accuracy and then improves steadily, whereas DAPO first falls from 17.8\% to 15.5\% and recovers only gradually.

By the final epoch, GCPO outperforms DAPO on all four datasets by 1.2, 4.9, 2.0, and 5.3 percentage points on AIME 2024, GSM8K, SVAMP, and TabMWP, respectively. The early emergence and persistence of these gains indicate that the failure-triggered auxiliary objective supplies useful supervision before the policy can reliably solve difficult prompts, improving both learning speed and final reasoning performance.

\section{Conclusion}

We introduce Group Contrastive Policy Optimization (GCPO), a hybrid post-training objective that complements on-policy GRPO with failure-triggered proximal contrastive supervision. Its central contribution is to formulate the offline golden trajectory as an empirical positive in a separate auxiliary objective rather than forcing it into the on-policy surrogate. This makes the sampling mismatch explicit: standard RLVR remains unchanged for informative groups, while all-wrong groups contrast verified reasoning against the policy's current failures. Our analysis explains how this hybrid design restores directional gradients exactly when group-relative advantages collapse, concentrates external supervision on difficult prompts, and transfers structured reasoning through prompt-specific positive--negative comparisons. It also clarifies that teacher-forced ratios provide proximal control rather than importance correction. These are mechanistic insights, not claims of unbiased off-policy estimation or global convergence. Experiments across two model scales and 10 core benchmarks demonstrate consistent gains in learning efficiency and final performance, while ablations support selective activation and positive--negative contrast. Additional case studies and more extensive results, including qualitative reasoning comparisons and complete training dynamics, will be provided in the appendix.

\section{Limitations}

A limitation of GCPO is that it requires the availability of complete and reliable golden trajectories, whose collection incurs additional computational or human annotation costs, and imperfect teacher reasoning may introduce noisy supervision. Its proximal contrastive term is an empirical auxiliary objective rather than an unbiased off-policy policy-gradient estimator; the teacher distribution is not modeled, and stronger off-policy corrections remain an open direction. Our evaluation is also limited to mathematical reasoning, two model scales, and one golden trajectory per prompt. Our setup further introduces a potential teacher--student alignment bias. Golden trajectories are generated by DeepSeek-R1, while both evaluated base models are DeepSeek-R1-distilled checkpoints, so their reasoning distributions may be closer than those of independently trained teacher and student models. We therefore have not established whether GCPO remains equally effective with a teacher from a different model family, which should be examined through cross-family transfer experiments.

\bibliography{aaai2027}

\end{document}